\documentclass[runningheads]{llncs}
\usepackage{graphicx}
\usepackage{makecell}

\usepackage{multirow}
\usepackage{array}
\usepackage{listings}
\usepackage{xspace}
\usepackage{mmstyle}
\usepackage{graphicx}
\usepackage{booktabs}
\usepackage{multirow}
\usepackage{tikz}
\usepackage{comment}
\usepackage{amsmath,amssymb} 
\usepackage{color}
\usepackage{tabularx}
\usepackage{adjustbox}
\usepackage{subcaption}
\usepackage[misc]{ifsym} 
\definecolor{linkcolor}{RGB}{255,0,0}
\definecolor{urlcolor}{RGB}{255,105,180}
\definecolor{citecolor}{RGB}{66,168,235}
\usepackage{hyperref}
\hypersetup{colorlinks=true,linkcolor=linkcolor,urlcolor=urlcolor,citecolor=citecolor}

\usepackage[accsupp]{axessibility}  

\begin{document}
\pagestyle{headings}
\mainmatter
\def\ECCVSubNumber{2087}  

\title{Fashionformer: A Simple, Effective and Unified Baseline for Human Fashion Segmentation and Recognition} 

\titlerunning{FashionFormer}
%
\author{Shilin Xu\inst{1,3*} \quad
Xiangtai Li\inst{1,3} \thanks{The first two authors contribute equally. The work was done at Sensetime Research. This research is also supported by the National Key Research and Development Program of China under Grant No.2020YFB2103402. We thank the computation resource provided by SenseTime Research.} \quad
Jingbo Wang\inst{2} \quad  \\
Guangliang Cheng\inst{3} \quad
Yunhai Tong\inst{1\textrm{\Letter}} \quad
Dacheng Tao\inst{4}}
\authorrunning{S. Xu and X. Li et al.}
\institute{\small Key Laboratory of Machine Perception, MOE, School of Artificial Intelligence, Peking University \\ \and CUHK-SenseTime Joint Lab, The Chinese University of Hong Kong \and SenseTime Research \and 
The University of Sydney \\
\email{lxtpku@pku.edu.cn, xushilin@stu.pku.edu.cn,chengguangliang@sensetime.com}}

\newcommand{\jingbo}[1]{{\color{blue} (jingbo: {#1})}}
\newcommand{\gl}[1]{{\color{red} (gl: {#1})}}
\maketitle

\begin{abstract}
Human fashion understanding is one crucial computer vision task since it has comprehensive information for real-world applications.
This focus on joint human fashion segmentation and attribute recognition. Contrary to the previous works that separately model each task as a multi-head prediction problem, our insight is to bridge these two tasks with one unified model via vision transformer modeling to benefit each task. In particular, we introduce the object query for segmentation and the attribute query for attribute prediction. Both queries and their corresponding features can be linked via mask prediction. Then we adopt a two-stream query learning framework to learn the decoupled query representations.We design a novel Multi-Layer Rendering module for attribute stream to explore more fine-grained features.  The decoder design shares the same spirit as DETR. Thus we name the proposed method \textit{Fahsionformer}. Extensive experiments on three human fashion datasets illustrate the effectiveness of our approach. In particular, our method with the same backbone achieve \textbf{relative 10\% improvements} than previous works in case of \textit{a joint metric (AP$^{\text{mask}}_{\text{IoU+F}_1}$) for both segmentation and attribute recognition}. To the best of our knowledge, we are the first unified end-to-end vision transformer framework for human fashion analysis. We hope this simple yet effective method can serve as a new flexible baseline for fashion analysis. Code will be available  \url{{https://github.com/xushilin1/FashionFormer}}.
\end{abstract}
\section{Introduction}
\label{sec:intro}
\begin{figure*}[t!]
	\centering
	\includegraphics[width=0.85\linewidth]{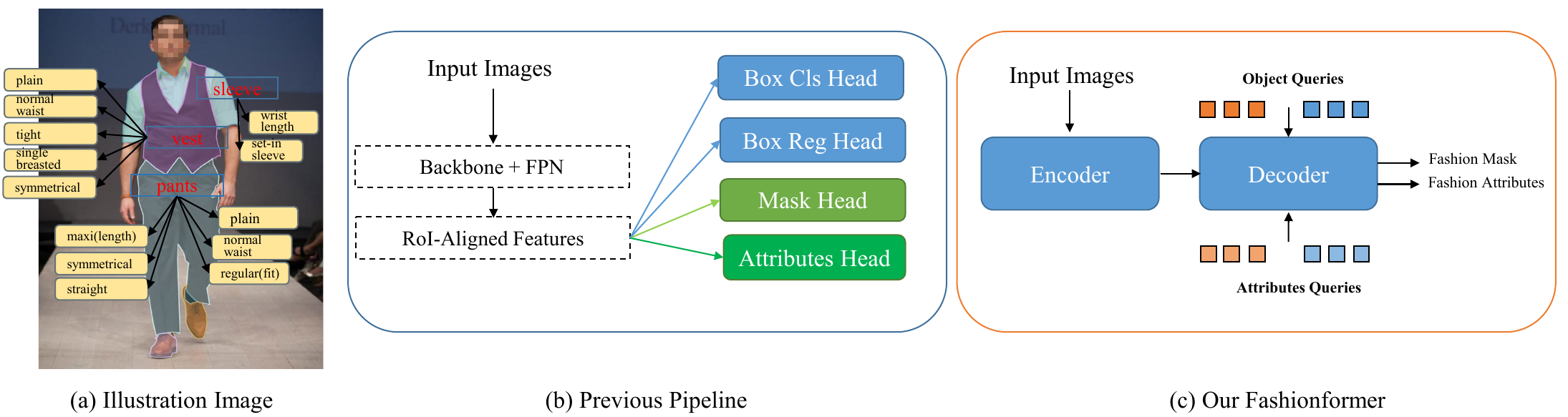}
	\caption{\small 
	Task introduction and method overview. (a) Illustration for human fashion segmentation and attribute recognition task. It requires the model to predict the masks, their labels (shown in red), and corresponding attributes (shown in yellow rectangles). (b) Overview of previous pipeline used in~\cite{jia2020fashionpedia}. It contains several independent task heads and solves each task separately. (c) The proposed Fashionformer pipeline. It is an encoder-decoder framework that takes object queries and attribute queries as inputs and directly outputs masks and attributes. Best view in color and zoom in.}
	\label{fig:teaser_01}
\end{figure*}

The capability of understanding human fashion is essential for numerous real-world applications, such as digital human modeling, AR/VR, and online business. Limited by the representation of human fashion, previous works~\cite{liuLQWTcvpr16DeepFashion,zheng_modanet,yamaguchi2014retrieving,yamaguchi2012parsing,yang2014clothing} mainly focus on fashion parts localization and ignore the fine-grained human attributes, for example, the description of clothes. To analyze such a richer context for human fashion, researchers~\cite{jia2020fashionpedia} begin to categorize human fashion attributes into different aspects (\eg plain, tight, normal waist) shown in Fig.~\ref{fig:teaser_01}(a). They solve this problem by extending the widely used instance segmentation framework~\cite{maskrcnn} for these attributes as a multi-head prediction task, as shown in Fig.~\ref{fig:teaser_01}(b). 



Although the Attribute-Mask R-CNN~\cite{jia2020fashionpedia} is a strong baseline for such fashion analysis task, there are several remaining critical issues. \textit{Firstly}, the relationship between instance segmentation and attribute recognition is not well explored and thus the fine-gained knowledge provided by these attributes for segmentation is unknown. \textit{Secondly}, adopting single RoI (Region of Interest) representation leads to incomplete and inferior results for fine-grained attributes, causing huge gaps between segmentation results and attribute recognition. \textit{Thirdly}, adopting box based detector limits the mask resolution and results in imprecise mask quality, which leads to uncomfortable experience for auto-segmentation when shopping. Thus, a better and unified baseline is needed for this track.

In recent years, Vision Transformers have been proven effective in feature representation learning, unification, and simplification of various downstream tasks~\cite{detr,peize2020sparse,VIS_TR,VIT,deit_vit}. In particular, Detection Transformer (DETR)~\cite{detr} adopts \textit{object queries} to model the detected objects. Meanwhile, several works~\cite{zhang2021knet,cheng2021mask2former} provide box-free segmentation solutions. Adopting query based design can naturally link segmentation and attribute recognition. More details can be obtained via corresponding masks. Moreover, as proven by ViT~\cite{VIT}, Vision Transformers have a huge capacity, which is helpful for the shopping industry since the data is easy to obtain. These findings motivate us to formulate a new solution for fashion analysis using transformers. 

\begin{figure*}[t]
	\centering
	\includegraphics[width=0.70\linewidth]{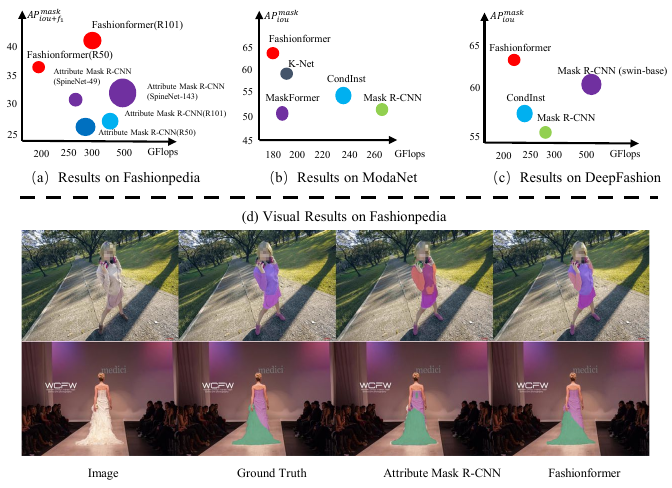}
	\caption{\small Comparison Results on (a) Fashionpedia, (b) ModaNet. and (c) DeepFashion. All models in (b) and (c) use ResNet50 as backbone. The GFlop is obtained via $1020 \times 1020 \times 3$ inputs. Our method achieves a significant gain with the fewer GFlops. The number of network parameters are indicated by the radius of circles. (d). Visual comparison results using ResNet-50 backbone. Best view in color.}
	\label{fig:teaser_02}
\end{figure*}

In this work, we present a simple, effective and unified baseline named Fashionformer for this task. As shown in Fig.~\ref{fig:teaser_01}(c), the entire framework is an encode-decoder framework in DETR style. Our key insights are using \textit{object queries} and \textit{attribute queries} to encode
instance-wised and attribute-wised information, respectively. Then we perform joint learning for both instance segmentation and attribute recognition. Joint learning leads to better instance segmentation results, since attribute labels constrain the scope of mask label. For example, for sleeve object, there will not exist attributes like plain or tight. In particular, we propose to match each object query into each attribute query via residual addition for implementation. This gives a good initialization of attributes and serves as the starting points for attribute recognition. Since each object query owns its corresponding mask, we name the query feature as grouped features via these corresponding masks. Then we perform updating object queries with the query features. This operation is implemented with dynamic convolution~\cite{zhang2021knet,peize2020sparse} and multi-head self-attention layers~\cite{vaswani2017attention} between query and query features iteratively. In the experiments, surprisingly, we find \textbf{\textit{joint}} learning with the object queries and attribute queries boosts the original segmentation results. 

Moreover, since both attribute recognition and instance segmentation have different goals, the former needs more fine-grained and discriminative representation for multi-label classification while the latter performs mask based multi-class classification. For attribute recognition, we propose a Multi-Layer Rendering (MLR) module to generate the attributes query features. The multi-level features are grouped via corresponding predicted masks. Such a module is lightweight and bridges the gap between instance segmentation and attribute recognition. Then we perform the same procedure as object query path via dynamic convolution and self-attention layers where we use attributes query feature to re-weight the attribute query. Note that both are \textit{independent}, which leads to a two-stream decoder.  

Our framework can support both CNN backbone~\cite{resnet} and Transformer backbone~\cite{liu2021swin} as encoder. Finally, the framework is motivated by the design of box-free architecture, and it can generate high resolution masks without box supervision or RoI-wised operation.Our framework can also adapt to previous fashion benchmarks by replacing the recognition branch with an extra classification head. Extensive experiments show that our approach achieves much better results than previous works~\cite{jia2020fashionpedia,zhang2021knet,cheng2021maskformer}. Fig.~\ref{fig:teaser_02} (d) shows our method achieves better segmentation results when comparing to Attribute Mask-RCNN. As shown in Fig.~\ref{fig:teaser_02}(a)-(c), Fashionformer achieves the new state-of-the-art results on three fashion benchmarks including Fahsionpedia~\cite{jia2020fashionpedia}, ModaNet~\cite{zheng_modanet} and Deepfashion~\cite{liuLQWTcvpr16DeepFashion} with \textit{less computation cost and parameters}. Moreover, \textit{The gap between attribute recognition and instance segmentation can be effectively narrowed through our approach by a significant performance margin to the state-of-the-art methods}. These results demonstrate that Fashionformer can serve as a new baseline for future research in fashion analysis. To the best of our knowledge, this is the first unified end-to-end model for this task.
\section{Related Work}
\label{sec:related_work}

\noindent
\textbf{Instance Segmentation.} This task aims to detect and segment each instance~\cite{dai2016instance,simultaneous_det_seg}. The two-stage pipeline Mask R-CNN like models~\cite{maskrcnn,msrcnn,htc,jia2020fashionpedia} first generate object proposals using Region Proposal Network (RPN)~\cite{ren2015faster} and then predict boxes and masks on each RoI where the advanced versions utilize more cues to enhance the mask representation. Several single-stage methods~\cite{tian2020conditional,chen2020blendmask,yolact-iccv2019,chen2019tensormask} achieve significant progress and comparable results with two-stage pipelines where they use single stage detector. SOLO~\cite{wang2019solo} treats instances into grid representation and then performs instance classification and segmentation in a decoupled manner. Meanwhile, there are several bottom-up approaches~\cite{neven2019instanceSeg,de2017semanticInstanceLoss} where each instance is grouped from the semantic segmentation
prediction~\cite{deeplabv3plus,zhou2022rethinking,li2020improving,xiangtl_gald}. Several works~\cite{QueryInst,dong2021solq} use object query to encode instance wised information. However, they still need object detector~\cite{zhu2020deformabledetr,peize2020sparse}. Instance Segmentation is one of the sub-task of our framework. Moreover, we show that joint learning of both segmentation and fine-grained attributes lead to better segmentation results. 

\noindent
\textbf{Related Human Segmentation.} 
Most works in this area focus on human part modeling, including instance part segmentation and semantic part segmentation. Several works~\cite{fang2018weakly,liu2018cross,wang2019CNIF,zhou2020cascaded} design specific methods for semantic part segmentation which are in category-level settings. Recent human segmentation methods focus on human instance part segmentation. There are two paradigms for this direction: \textit{top-down} pipelines~\cite{li2017holistic,yang2019parsing,ruan2019devil,ji2019learning,yang2020eccv} and \textit{bottom-up} pipelines~\cite{gong2018instance,li2017multiple,zhao2018understanding,zhou2021differentiable}. Our framework solves the \textit{instance level} segmentation with \textit{fine-grained attributes recognition}, which is much more challenging. 

\noindent
\textbf{Fine-grained Image Recognition.} Lots of works~\cite{fu2017look,zheng2017learning} use the localization classification subnetwork to highlight finer feature regions and then achieves better classification results. Various approaches are proposed including attention~\cite{zhang2019learning}, extra models~\cite{zhang2014part,zhang2016spda} and deep filters~\cite{xiao2015application,wang2018learning}. Meanwhile, there are several works learning the end-to-end feature encoding via specific loss~\cite{sun2018multi} or high-order feature interactions~\cite{lin2015bilinear}. Our methods adopt object query and object mask as extra spatial cues to grasp multiscale image features, which lead to better fine-grained recognition results. 

\noindent
\textbf{Transformer in Computer Vision.} There are two research directions for vision transformer. The first is to replace CNN as a feature extractor. Compared with CNN, vision transformers~\cite{VIT,liu2021swin,deit_vit} have more advantages in modeling long-range relation among the patch features. Moreover, they show better performance and higher capacity among the downstream tasks. The second design is to use the object query representation. DETR~\cite{detr} models the object detection task as a set prediction problem with object queries. The following works~\cite{zhu2020deformabledetr,zhang2021knet,cheng2021mask2former} explore the locality of the learning process to improve the performance of DETR. Meanwhile, there are also several works using object queries to solve more complex tasks~\cite{zhou2022transvod,yuan2021polyphonicformer,panopticpartformer,VIS_TR,video_knet}.
Our work is inspired by these works, which also uses a transformer architecture to unify and simplify human fashion task. Thus, our main contributions lie in the second part of vision transformer. However, our method can be adapted with vision transformer backbone~\cite{liu2021swin}.
\section{Method}
\label{sec:method}
\noindent
In this section, we will first review the related works using object query and present our method's key insights and motivations. Then we briefly describe our Fashionformer and detail the design of our proposed two-stream query learning framework for both object query and attribute query. Finally, we describe the training and inference procedures. 

\subsection{Overview of Previous Work}

\noindent
\textbf{Query based models.} DETR~\cite{detr} firstly introduces the concept of object query which is used as the input of transformer decoder to build one by one mapping on objects in the scene. Various approaches in different tasks~\cite{VIS_TR,zhu2020deformabledetr,cheng2021maskformer} prove the effectiveness of such design including segmentation and detection. The MaskFormer~\cite{cheng2021maskformer,cheng2021mask2former} shows that the pure mask based classification can solve both semantic and instance level segmentation problems. Meanwhile, several works~\cite{zhang2021knet,peize2020sparse} show that adopting self-attention on proposal level features with their corresponding object queries can also achieve robust performance. For both aspects, the critical component is the design of object query and interaction with feature maps from backbone. 

\noindent
\textbf{Motivation of our method.} As stated in Sec.~\ref{sec:intro}, recent works~\cite{jia2020fashionpedia,liuLQWTcvpr16DeepFashion} on fashion analysis have several shortcomings. In this work, we seek a new baseline for unified fashion analysis to solve both segmentation and attribute recognition jointly. Motivated by the recent process of query based models, a natural question emerges: \textit{can a unified model improves both instance segmentation and attribute recognition, despite the goals of these two tasks being different?} Such significant gaps between two tasks have already existed in the results of previous work~\cite{jia2020fashionpedia}. Our insights are: Firstly, we add an extra attribute query with a residual like learning with object query to balance the conflicts of two tasks. Secondly, since both queries share the same instance masks, we decouple the query feature learning via designing specific modules for each task, which results in a two-stream decoder. We detail the design in the following sections.  

\begin{figure*}[t!]
	\centering
	\includegraphics[width=0.80\linewidth]{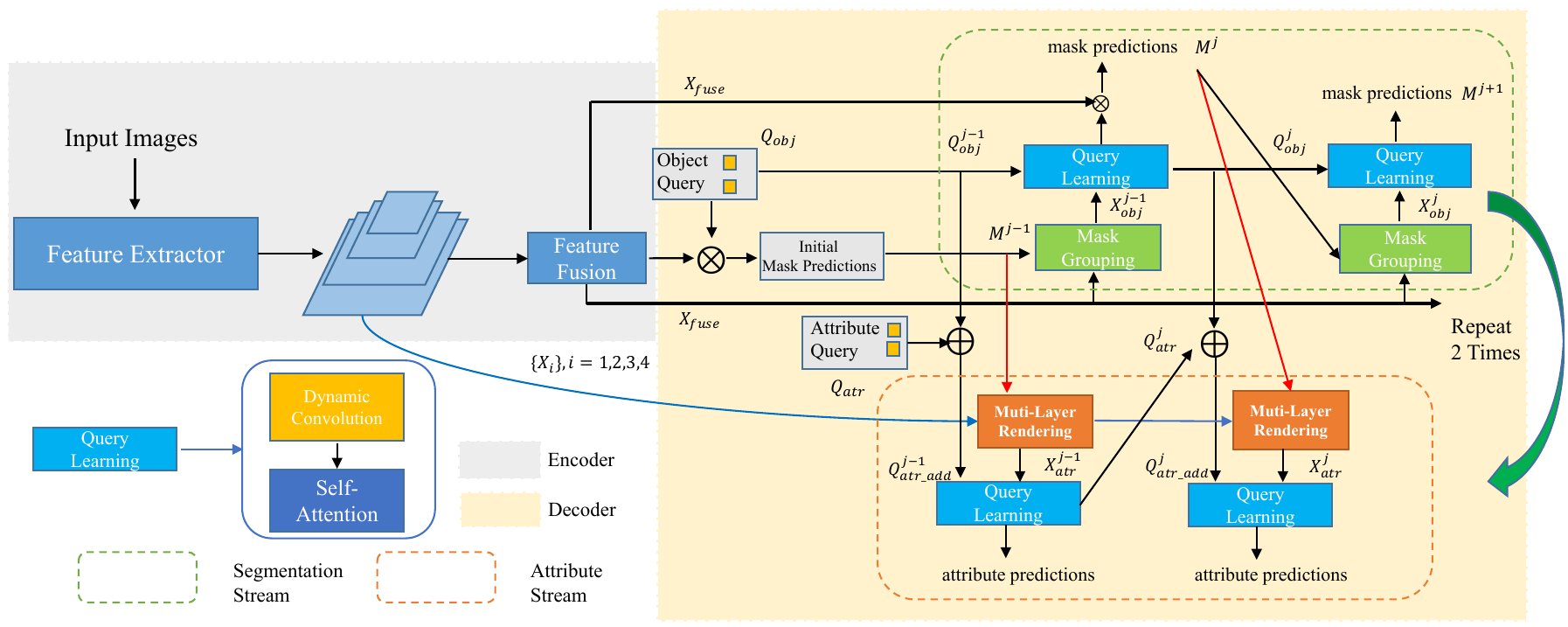}
	\caption{\small Fashionformer contains three parts. (1) a feature extractor to extract multiscale features and then fuse these feature into one high resolution. (2) segmentation stream to generate segmentation masks and labels. It takes the object queries and initial mask as inputs and directly outputs refined mask and object queries interactively. (3) attribute stream to obtain the attribute predictions. It takes attribute queries, mask predictions and multiscale features as inputs and outputs the attribute prediction.}
	\label{fig:methods}
\end{figure*}

\subsection{Fashionformer as a Unified Baseline} 

\noindent
\textbf{Feature extractor.} We first extract image features for each input image using a feature extractor. It contains a backbone network (Convolution Network~\cite{resnet} or Vision Transformer~\cite{liu2021swin}) with Feature Pyramid Network~\cite{fpn} as neck. This results in a set of multiscale features $\{X_i,..\}$ where $i={1,2,3,4}$ are indexes of different scales. Then we sum up all the feature pyramids into one high resolution feature map $X_{fuse}$ which follows the design of~\cite{kirillov2019panopticfpn}. Moreover, we follow the common design of DETR-like models~\cite{detr,tian2020conditional,zhang2021knet} to add position embeddings on the $X_{fuse}$. We omit this operation for simple illustration. 

Meanwhile, we keep the multiscale features $\{X_i,..\}$ for further usage. This process is shown in the gray region of Fig.~\ref{fig:methods}. We denote the \textit{object query feature} as the corresponding feature from feature extractor (shown in black line, $X_{fuse}$) via mask grouping, which is used in segmentation stream. We denote the \textit{attribute query features} as corresponding feature generated by Multi-Layer Rending which is used in attribute stream. We detail the process further. 

\noindent
\textbf{Initial object query.} Following previous works~\cite{zhang2021knet,peize2020sparse}, the initial weights of object queries are directly obtained from the weights of initial decoder prediction. We adopt one $1\times1$ convolution layer to obtain the initial instance masks $M_{0}$. The kernel weights are the query weights. Such initialization shows better convergence. The object queries $Q_{obj}$ are in $N \times d$ where $N$ denotes the query number ($N=100$ by default) and $d$ is the hidden dimension ($d=256$ by default, which is the same as DETR~\cite{detr}).

\noindent
\textbf{two-stream decoder architecture.} As shown in the yellow region of the Fig.~\ref{fig:methods}, we adopt a two-stream design for both segmentation and attribute recognition. In particular, the attribute recognition module is appended to the segmentation module. Since most detection transformers~\cite{detr,zhang2021knet,cheng2021maskformer} have cascaded decoder design to converge, we keep the same strategy via repeating the both modules into a cascaded decoder. In each step, the segmentation module directly outputs the binary segmentation masks and mask classification results while the recognition module outputs the attribute probability maps. We use index $j$ to represent step number. All the results are supervised with specific loss functions during the training.

\subsection{Linking Object Query and Attributes Query}

\noindent
\textbf{Adding extra attribute query.} Rather than directly learning the attribute query from the object query via a multi-layer fully connected network, we present an extra attribute query design where we initialize a learnable embedding $Q_{atr}$ with the same shape as $Q_{obj}$. For each step $j$ in the two-stream decoder, we add both queries together as the new attribute query:

\begin{equation}
     {Q}_{atr\_add}^{j-1} = Q_{atr}^{j-1} + Q_{obj}^{j-1}. 
     \label{attribute_addtion}
\end{equation}
When $j=1$ for the initial stage ($j \in \{1,2,3\}$), we denote ${Q}_{atr} = Q_{atr}^{0}$, $Q_{obj} = Q_{obj}^{0}$ in the following sections. 

\noindent
\textbf{Multi-Layer Rendering (MLR).} This module takes previous multi-level features $\{X_{i}, ...\}, i=1,2,3,4$ (\textcolor{blue}{blue arrows} in Fig.~\ref{fig:methods}) and mask predictions $M^j-1$ (\textcolor{red}{red arrows} in Fig.~\ref{fig:methods}) from the previous stage as the inputs. It outputs a refined attribute query feature $X^j_{atr}$. We first resize the $M^{j-1}$ into different scales ($i=1,2,3,4$) and then perform multi-level grouping with multi-level features $X_{i}$ as following:
\begin{equation}
    X^{j}_{i,atr} = {\sum_u^{W_{i}}\sum_v^{H_{i}} M^{j-1}(u, v, i) \cdot X_{i}(u, v), i \in \{1,2,3,4\}, }
 \label{equ:grouping_mvl}
\end{equation}
where $j$ is the interaction number, $W_{i}$ and $H_{i}$ are the height and width of the corresponding features $X_{i}$, $u$ and $v$ are the spatial index of features. Then we obtain a set of features $X^{j}_{i,atr}$ with the same shape of attribute queries.
Then we adopt one MLP-like to fuse these queries where it contains one fully connected layer, one Layer Norm~\cite{vaswani2017attention} and one relu activation layer as:
\begin{equation}
    X^j_{atr} = MLP(Concat({X^{1,j}_{atr},..X^{i,j}_{atr}})), i \in \{1,2,3,4\} .
 \label{equ:rendering}
\end{equation}
This operation is shown in \textcolor{orange}{orange nodes} in Fig.~\ref{fig:methods}. The final refined attribute query feature $X^j_{atr}$ will be the input of attribute stream for attribute query learning. In Sec.~\ref{sec:exp}, we find this module significantly improves the attribute recognition results.

\noindent
\textbf{Decoupled query learning.} For object query feature, we mainly follow the design of previous works~\cite{zhang2021knet,peize2020sparse}. The object query feature $X_{obj}^{j-1}$ is obtained from the mask-based grouping from the previous mask prediction and feature map $X_{fuse}$ (Mask Grouping in \textcolor{green}{green nodes}) where:
\begin{equation}
    X_{obj}^{j-1} = \sum_u^W\sum_v^H M^{j-1}(u, v) \cdot X_{fuse}(u, v).
\end{equation}

Then, following~\cite{tian2020conditional,zhang2021knet,peize2020sparse}, we perform a Dynamic Convolution (DC) to refine input queries $Q_{obj}^{j-1}$ with the object query features $X_{obj}^j$. The motivation of DC is to encode instance-wise features into each object query for fast convergence. This is \textbf{not} our contribution.
\begin{equation}
    {Q}_{obj}^{j-1} = DC(X_{obj}^{j}, Q_{obj}^{j-1}),
 \label{equ:dynamic_obj}
\end{equation}
where the dynamic convolution uses the query features $X_{obj}^j$ to re-weight input queries $Q_{obj}^{j-1}$. To be more specific, $DynamicConv$ uses object query features $X_{obj}^{j}$ to generate gating parameters via MLP and multiply back to the original query input $Q_{obj}^{j-1}$. For attribute branch, we adopt the same procedure:  
\begin{equation}
    {Q}_{atr}^{j-1} = DC(X_{atr}^{j}, Q_{atr\_add}^{j-1}),
 \label{equ:dynamic_atr}
\end{equation}
where ${X}_{atr}^{j}$ is the \textit{attribute query feature} from the MLR. Note that both $DC$ operations have independent parameters and learned in a separate manner. In this way, both streams are decoupled and lead to better attribute recognition results. 

We adopt the same design~\cite{zhang2021knet} by learning gating functions to update the refined queries including both object queries and attributes queries. The $DC$ operation is shown as follows:
\begin{equation}
    \hat{Q}_{u}^{j-1} = gate_{x}(X_{u}^{j})X_{u}^{j} + gate_{q}(X_{u}^{j}) Q_{u}^{j-1}, u \in \{obj, atr \},
\end{equation}
where $gate$ is implemented with a fully connected (FC) layers followed by LayerNorm (LN), and a sigmoid layer. We adopt two different gate functions, including $gate_{x}$ and $gate_{q}$. The former is to weight the query features, while the latter is to weight corresponding queries. After that, one self-attention layer with feed forward layers~\cite{vaswani2017attention,wang2020maxDeeplab} is used to learn the correspondence among each query. This operation leads to the full correlation among queries shown as follows:
\begin{equation}
    Q_{u}^{j} = FFN(MHSA(\hat{Q}_{u}^{j-1}) + \hat{Q}_{u}^{j-1}), u \in \{obj, atr\},
 \label{equ:selfattention}
\end{equation}
where $MHSA$ means Multi Head Self Attention, $FFN$ is the Feed Forward Network that is commonly used in current vision transformers~\cite{detr,VIT}. The output refined query sets contain $Q_{obj}^{j}$ and $Q_{atr}^{j}$. Both query learning works independently, shown in \textcolor{blue}{blue nodes} in Fig.~\ref{fig:methods}. The main reason for using independent query learning is the optimization goal. The attribute query for attribute prediction is \textit{a multi-label classification problem}, while the object query for segmentation is \textit{a single label classification problem}. 

\noindent
\textbf{Generating mask and attributes prediction.}
Finally, the refined masks are obtained via dot product between the refined queries $Q_{obj}^j$ and the input features $X_{fuse}$. For mask classification, we adopt two feed forward layers on $Q_{obj}^{j}$ and directly output the class scores. For mask segmentation, we also adopt two feed forward layers on $Q_{obj}^{j}$ and then we perform the inner product between learned queries and features $X_{fuse}$ to generate stage $i$ object masks. These masks will be used for the next step $j+1$. 

For attribute prediction, we also adopt several feed forward layers on $Q_{atr}^{j}$ and use the outputs with a sigmoid activation function as final output. The process of these equations will be repeated several times. The iteration number is set to 3 by default. All the inter mask predictions are refined and optimized during the training.

\subsection{Training and Inference}

\noindent
\textbf{Loss functions.} We mainly follow the design of previous works~\cite{cheng2021maskformer,wang2020maxDeeplab,detr} to use bipartite matching as a cost by considering both mask and classification results. After the bipartite matching, we apply a loss jointly considering mask prediction and classification for object queries. In particular, we apply focal loss~\cite{focal_loss} on both classification and mask prediction. We also adopt dice loss~\cite{dice_loss} on mask predictions. For attribute queries, we follow the default design of the Attirbute Mask R-CNN and the prediction is supervised in one hot format, which is the default setting of multi-label classification. It is a binary multi-label classification loss. The total loss can be written as follows:
\begin{equation}
    \mathcal{L}_{j} =  \lambda_{cls} \cdot \mathcal{L}_{cls} + \lambda_{mask} \cdot \mathcal{L}_{mask} +\lambda_{atr} \cdot \mathcal{L}_{atr}.
\end{equation}
Note that the losses are applied to each stage $\mathcal{L}_{final} = \sum_{j}^N\mathcal{L}_j,$ where $N$ is the total stages applied to the framework. We adopt $N = 3$ and all $\lambda$s are set to 1 by default.

\noindent
\textbf{Inference.} We directly get the output masks from the corresponding queries according to their sorted scores, which are obtained from the mask classification branch. Since object queries and attribute queries are in the one-by-one mapping manner. We obtain the attribute predictions with 0.5 threshold via a sigmoid function from attribute prediction.
\section{Experiment}
\label{sec:exp}
%
%

\subsection{Settings}
\noindent
\textbf{Dataset.} We carry out our experiments on Fashionpedia~\cite{jia2020fashionpedia}, ModaNet ~\cite{zheng_modanet} and DeepFashion~\cite{liuLQWTcvpr16DeepFashion} dataset. Fashionpedia is a new challenging fashion dataset which contains 45,632 images for training and 1,158 images for validation. ModaNet contains 52,254 images for training. Since the online server of ModaNet is closed, we randomly sample 4,000 images from the training dataset for testing. DeepFashion datasets have 6,817 images for training and 3,112 images for testing.  For both ModaNet and DeepFashion datasets, we retrain the several representative baselines~\cite{maskrcnn,tian2020conditional,zhang2021knet} for fair comparison. We use the challenging Fashionpedia dataset for ablation and analysis, since it contains both segmentation and attribute labels. We only report the instance segmentation results on the two remaining datasets.

\noindent
\textbf{Implementation details.} We implement our models in PyTorch~\cite{pytorch_paper} with MMDetection toolbox~\cite{chen2019mmdetection}. We use the distributed training framework with 8 GPUs. \textit{For Fashionpedia dataset}, we adopt similar settings as the original work. We use large scale jittering that resizes an image to a random ratio between $[0.5, 2.0]$ of the target input image size. ResNet-50, ResNet-101~\cite{resnet} and Swin-Transformer~\cite{liu2021swin} are used as the backbone network and the remaining layers use Xavier~\cite{xavier_init} initialization. The optimizer is AdamW~\cite{ADAMW} with weight decay 0.0001. The training batch size is set to 16 and all models are trained with 8 GPUs. We adopt 12 epochs training (one-third of 1 $\times$ schedule by default, 12 epochs in original size.) for ablation studies. We empirically found more training epochs leads to better results. All the models adopt the single scale inference. \textit{For both ModaNet and DeepFashion dataset}, we follow the same setting from COCO~\cite{coco_dataset} for fair comparison. Note that since both datasets have no attribute labels, we treat the mask classification labels as one attribute without the modification of network architecture. We present more implementation details in appendix. 


\noindent
\textbf{Metrics.} Following previous works, we adopt mask based mean average precision (AP$^{\text{mask}}_{\text{IoU}}$) which is the default COCO setting for instance segmentation evaluation. Moreover, we also adopt mAP that also consider F1-score (AP$^{\text{mask}}_{\text{IoU+F}_1}$) for joint evaluating segmentation and attribute recognition. This is the \textbf{main metric} for comparison. The GFlops are obtained with $3\times1020\times1020$ inputs following the original work~\cite{jia2020fashionpedia}. Moreover, we also report the gap $G$ between AP$^{\text{mask}}_{\text{IoU}}$ and AP$^{\text{mask}}_{\text{IoU+F}_1}$ since the latter is more challenging.

\begin{table*}[!t]
	\centering
    \scalebox{0.70}{\subfloat[Effect of each component.
    ]{
        \label{tab:effect_component}
	    \begin{tabularx}{0.50\textwidth}{c c c c c c c} 
		        				\toprule[0.15em]
    		RD & DC  & MLR & N=1 & I=3 & AP$^{\text{mask}}_{\text{IoU}}$ & AP$^{\text{mask}}_{\text{IoU+F}_1}$  \\
            \midrule[0.15em]
    		\checkmark & \checkmark & \checkmark &  - & \checkmark & \textbf{33.2} & \textbf{29.2} \\
    		- & \checkmark & \checkmark & - & \checkmark &  33.0 & 27.0 \\ 
    		\checkmark & - & \checkmark  & - & \checkmark & 32.1 & 27.5 \\ 
    		\checkmark & \checkmark & - & - & \checkmark & 33.0 & 26.5 \\ 
    		\checkmark & \checkmark & \checkmark & \checkmark & - & 29.1 & 25.8 \\ 
    		- & \checkmark & - & - & \checkmark & 30.1  & - \\ 
        	\bottomrule[0.1em]
	    \end{tabularx}
    }} \hfill
    \scalebox{0.70}{
    \subfloat[Ablation on design of MLR. NL: number of layers in MLR.]{
        \label{tab:ablation_MLR}
		\begin{tabularx}{0.30\textwidth}{c c c} 
			\toprule[0.15em]
		  Setting & AP$^{\text{mask}}_{\text{IoU}}$ & AP$^{\text{mask}}_{\text{IoU+F}_1}$ \\
			\midrule[0.15em]
    		  NL=1 & 32.2 & 27.5 \\
              NL=2 & 33.1 & 28.6 \\
              NL=4 & 33.2 & \textbf{29.2} \\
			\bottomrule[0.1em]
		\end{tabularx}
    }} \hfill
    \scalebox{0.60}{
    \subfloat[Decoupled query learning. The Shared Query use the same object query via a MLP to generate attribute query.]{
        \label{tab:decoupled_ql}
		\begin{tabularx}{0.50\textwidth}{c c c c} 
			\toprule[0.20em]
			Setting & AP$^{\text{mask}}_{\text{IoU}}$ & AP$^{\text{mask}}_{\text{IoU+F}_1}$  & Param(M) \\
			\midrule[0.15em]
			Shared Query  & 33.4 & 27.6 & 36.2\\ 
            Individual Query  & 33.3 & \textbf{29.1} & 37.7 \\ 
			\bottomrule[0.1em]
		\end{tabularx}
    }} \hfill
	\caption{\small Ablation studies and analysis on Fahsionpedia dataset set with ResNet50 as backbone. DC: Dynamic Convolution. RD: Residual Addition. MLR: Multi-Layer Rendering.
	N: Number of decoder layers.}
\end{table*}

\begin{table*}[!t]
    \footnotesize
	\centering
    \scalebox{0.85}{
    \subfloat[Comparison on recent box based approaches.]{
        \label{tab:compare_box_based}
	    \begin{tabularx}{0.45\textwidth}{c c c } 
		        				\toprule[0.15em]
    		 Method  & backbone & AP$^{\text{mask}}_{\text{IoU}}$   \\
    		\toprule[0.15em]
        		Mask-RCNN~\cite{maskrcnn} & ResNet50  & 30.3 \\
    		CondInst~\cite{tian2020conditional} & ResNet50 & 26.6 \\
    		QueryInst~\cite{QueryInst} & ResNet50 & 31.7 \\
    		Ours & ResNet50  &  \textbf{33.2} \\
        	\bottomrule[0.1em]
	    \end{tabularx}
    }} \hfill
    \scalebox{0.85}{
    \subfloat[Comparison on recent Transformer based models. 12e means 12 epochs training.]{
        \label{tab:compare_transformer_like_model}
	    \begin{tabularx}{0.50\textwidth}{c c c c} 
		  \toprule[0.15em]
    	    Method  & backbone & AP$^{\text{mask}}_{\text{IoU}}$ & schedule   \\
    		\toprule[0.15em]
        	K-Net~\cite{zhang2021knet} & ResNet50 & 30.3 & 12e	\\
        	MaskFormer~\cite{cheng2021maskformer}  & ResNet50 & 31.4 & 12e	 \\
        	Our & ResNet50 & \textbf{33.2} & 12e \\
        	\hline
        	K-Net~\cite{zhang2021knet} & Swin-b & 46.4 & 3$\times$ \\
        	Our & Swin-b &  \textbf{49.5} & 3$\times$ \\
        	\bottomrule[0.1em]
	    \end{tabularx}
    }} \hfill
    
	\caption{\small Comparison with recent representative methods on Fashionpedia.}
\end{table*}

\subsection{Ablation Studies.}

\noindent
\textbf{Effectiveness analysis of each component.} In Tab.~\ref{tab:effect_component}, we explore the effectiveness of each component in our Fashsionformer. Compared with instance segmentation baseline (the last row), our method (the first row) achieves significant improvements on instance segmentation (3.1\%). This indicates the effectiveness of our framework that the joint learning leads to better segmentation results. It means fine-grained attribute recognition can improve the fashion classification. Removing RD leads to inferior results on attribute recognition (about 2.0\% drop) which means the mask classification conflicts with attribute recognition. Removing DC also leads to bad results since the model does not converge which is also observed in previous works~\cite{peize2020sparse,zhang2021knet}. Moreover, from the first row and the fourth row, we find that adding MLR leads to a significant improvement (2.7\%) on attribute recognition \textit{without} hurting the performance of instance segmentation. MLR explores the multi-level features to make the attribute query features. This leads to a more distinctive query representation. Decreasing the number of decoder layers leads to a significant drop (from 33.2\% to 29.1 \%) which means more steps are needed for the decoder. However, adding more steps in decoder will not increase performance ($I=4$, 33.3\%). Details can be found in the appendix. 

\noindent
\textbf{Ablation on design of MLR module.} In Tab.~\ref{tab:ablation_MLR}, we explore the layer number that affects the final attribute recognition performance.
We increase the number in a top-down manner (low resolution to high resolution). As shown in that table, we find increasing the number of layers can lead to better results. This proves that better attribute recognition needs multi-level feature representation. Moreover, it only takes \textit{3\% GFlops} increase when appending the MLR. 

\noindent
\textbf{Ablation on design of decoupled query learning.} We also explore the decoupled query learning in Equ.~\ref{equ:dynamic_obj}, Equ.~\ref{equ:dynamic_atr} and Equ.~\ref{equ:selfattention}. We find using independent dynamic convolution and self-attention modules can lead to better results
than shared query learning where both Equ.~\ref{equ:dynamic_obj} and Equ.~\ref{equ:dynamic_atr} share \textit{the same parameters}. This verifies our design of two-stream architecture and shows that both tasks need decoupled query feature learning to avoid the conflicts.

\subsection{Comparison with State-of-the-art Methods}

\begin{table}[t]
\begin{center}
\scalebox{0.75}{
\begin{tabular}{c | c | c | c | c | c | c | c }
\Xhline{1.0pt}
\textbf{method} & \textbf{backbone} & \textbf{schedule} & \textbf{GFlops} & \textbf{params(M)} & \textbf{AP$^{\text{mask}}_{\text{IoU}}$  $\uparrow$} & \textbf{AP$^{\text{mask}}_{\text{IoU+F}_1}$  $\uparrow$ } & \textbf{$G$   $\downarrow$ }\\
\Xhline{1.0pt}
\multirow{3}{*}{\makecell{Attirbute-Mask R-CNN}} & \multirow{3}{*}{R50-FPN}  & 1$\times$ & \multirow{3}{*}{296.7} &  \multirow{3}{*}{46.4} & 34.3 & 25.5  & 8.8 \\ 
& & 2$\times$ & & & 38.1 & 28.5 & 9.6 \\
& & 3$\times$ & & & 39.2 & 29.5 & 9.7\\
\hline
\multirow{3}{*}{Attirbute-Mask R-CNN} & \multirow{3}{*}{R101-FPN} & 1$\times$ & \multirow{3}{*}{374.3} & \multirow{3}{*}{65.4} & 36.7 & 27.6 & 9.1 \\ 
& & 2$\times$ & & & 39.2 & 29.8 & 9.4 \\
& & 3$\times$ & & & 40.7 & 31.4 & 9.3\\

\hline
\multirow{3}{*}{Attirbute-Mask R-CNN} & SpineNet-49 & \multirow{3}{*}{6$\times$} & 267.2 & 40.8 & 39.6 & 31.4 & 8.2 \\
& SpineNet-96 &                            & 314.0 & 55.2 & 41.2 & 31.8 & 9.4 \\
& SpineNet-143 &                           & 498.0 & 79.2 & 43.1 & 33.3 & 9.8 \\
\Xhline{1.0pt}
\multirow{2}{*}{Fashionformer}& \multirow{2}{*}{R50-FPN} & 1$\times$ & \multirow{2}{*}{198.0} & \multirow{2}{*}{37.7} & \textbf{40.3} & \textbf{36.6} &  3.7 \\
& & 3$\times$ &  &  & \textbf{42.5} & \textbf{39.4} & 3.1 \\ 
\hline
\multirow{2}{*}{Fashionformer} &  \multirow{2}{*}{R101-FPN} & 1$\times$ & \multirow{2}{*}{275.7} & \multirow{2}{*}{56.6} & \textbf{43.2} & \textbf{40.5} &  \textbf{2.7} \\
& & 3$\times$ &  &  & \textbf{45.6} & \textbf{42.8} & \textbf{2.8} \\ 
\hline
Attirbute-Mask R-CNN & Swin-b & 3$\times$ & 508.3 & 107.3 & 47.5 & 40.6 & 6.9 \\
Fashionformer & Swin-b & 3$\times$ & 442.5 & 100.6 & \textbf{49.5} & \textbf{46.5} & 3.0 \\
\Xhline{1.0pt}
\end{tabular}}
\end{center}
\caption{\small Benchmark results on Fashionpedia. We report  AP$^{\text{mask}}_{\text{IoU}}$ , AP$^{\text{mask}}_{\text{IoU+F}_1}$ (main metric) and their gaps $G$. Swin-b means Swin-base~\cite{liu2021swin}. Our methods achieve significant improvement over these three metrics.}
\label{tab:fashionpedia_res}
\end{table}

\begin{table*}[!t]
    \footnotesize
	\centering
    \scalebox{0.75}{
    \subfloat[Results on ModaNet dataset.]{
        \label{tab:modanet_res}
	    \begin{tabularx}{0.50\textwidth}{c c c c} 
		        				\toprule[0.15em]
    		 Method  & backbone & AP$^{\text{mask}}_{\text{IoU}}$ & GFlops \\
    		\toprule[0.15em]
    	    Mask R-CNN~\cite{maskrcnn} & ResNet50 & 51.8 &  264.8\\ 
    	    Mask R-CNN~\cite{maskrcnn} & Swin-b & 54.8 &  508.8\\
    	    CondInst~\cite{tian2020conditional} & ResNet50 & 54.9 & 234.2\\
    	    K-Net~\cite{zhang2021knet} &  ResNet50  & 57.9 & 198.5\\
    	    QueryInst~\cite{QueryInst} & ResNet50 & 53.6 & 469.2 \\
    	    MaskFormer~\cite{cheng2021maskformer} & ResNet50 & 51.5 & 191.5 \\
            \hline
            Fashionformer & ResNet50 & 62.5 &  198.0 \\
        	\bottomrule[0.1em]
	    \end{tabularx}
    }} \hfill
    \scalebox{0.75}{
    \subfloat[Results on DeepFashion dataset.]{
        \label{tab:deepfashion_res}
	    \begin{tabularx}{0.50\textwidth}{c c c c} 
		  \toprule[0.15em]
    	    Method  & backbone & AP$^{\text{mask}}_{\text{IoU}}$ & GFlops \\
    		\toprule[0.15em]
    	    Mask R-CNN~\cite{maskrcnn} & ResNet50 & 56.9 & 264.8\\ 
    	    Mask R-CNN~\cite{maskrcnn}& Swin-b & 61.8  & 508.8\\
    	    CondInst~\cite{tian2020conditional} & ResNet50 & 58.6 & 234.2\\
    	    K-Net~\cite{zhang2021knet} &  ResNet50  & 62.0 & 198.5\\
    	    QueryInst~\cite{QueryInst} & ResNet50 & 61.4 & 469.2 \\
            \hline
            Fashionformer &  ResNet50 & 64.4 & 198.0\\
        	\bottomrule[0.1em]
	    \end{tabularx}
    }} \hfill
	\caption{\small Benchmark results on ModaNet~\cite{zheng_modanet} (a) and DeepFashion~\cite{liuLQWTcvpr16DeepFashion} (b).}
\end{table*}

\noindent
\textbf{Comparison with box-based approaches on Fashionpedia.}  In Tab.~\ref{tab:compare_box_based}, we compare several representative results on Fahsionpedia dataset using 12 epochs training. Our method achieves better result than recent instance segmentation methods including dynamic convolution based~\cite{tian2020conditional}, query based approaches~\cite{QueryInst}. This proves our query design is a better choice for fashion analysis.

\noindent
\textbf{Comparison with transformer based models on Fashionpedia.}
We also compare several recent pure mask based approaches in Tab.~\ref{tab:compare_transformer_like_model}. All the methods use the same codebase with the same setting. Our method outperforms previous works~\cite{zhang2021knet,cheng2021maskformer} by a large margin under various settings (both ResNet and Swin Transformer).

\noindent
\textbf{Results on Fashionpedia benchmark.} In Tab.~\ref{tab:fashionpedia_res}, we compare our model results with previous state-of-the-art approaches. Models that use the Swin Transformer are re-trained in our settings. We compare our methods with Attribute-Mask R-CNN in the different settings. In particular, using ResNet50 backbone, our method outperforms it by 6\% mAP$^{\text{mask}}_\text{IoU}$  and nearly 10\% mAP$^{\text{mask}}_{\text{IoU+F}_1}$. The gap $G$ is decreased nearly 50\%. When adopting the ResNet101 backbone, our method still achieves nearly 4\% mAP$^{\text{mask}}_{\text{IoU}}$ gain and 10\% mAP$^{\text{mask}}_{\text{IoU+F}_1}$ gain. It even outperforms the more complex backbone SpineNet~\cite{du2020spinenet} with more training iterations. Finally, we equip our method with the Swin-base backbone, it also outperforms the Attirbute-Mask RCNN by 5.7\% mAP$^{\text{mask}}_\text{IoU}$ and achieves the best performance. In the above three metrics, the improvements on mAP$^{\text{mask}}_{\text{IoU+F}_1}$ are more significant which means our model can achieve much better attribute recognition. This indicates the effectiveness of our framework including two stream decoder design and the proposed Multi-Layer Rendering module.

\noindent
\textbf{Results on ModaNet benchmark.} We further report results on the ModaNet benchmark in Tab.~\ref{tab:modanet_res}. Our Fashionformer achieves the best performance among these works with less GFlops. Note that our work with ResNet50 backbone even outperforms Mask-RCNN with Swin-base backbone~\cite{liu2021swin}. This indicates the effectiveness of our approach.

\noindent
\textbf{Results on DeepFashion benchmark.} In Tab.~\ref{tab:deepfashion_res}, we compare our methods with recent representative works~\cite{cheng2021maskformer,zhang2021knet}. Our works also achieve state-of-the-art results using ResNet50 backbone which further shows the generalization ability of our approach.

\subsection{Analysis and Visualization}

\begin{figure*}[t!]
	\centering
	\includegraphics[width=0.75\linewidth]{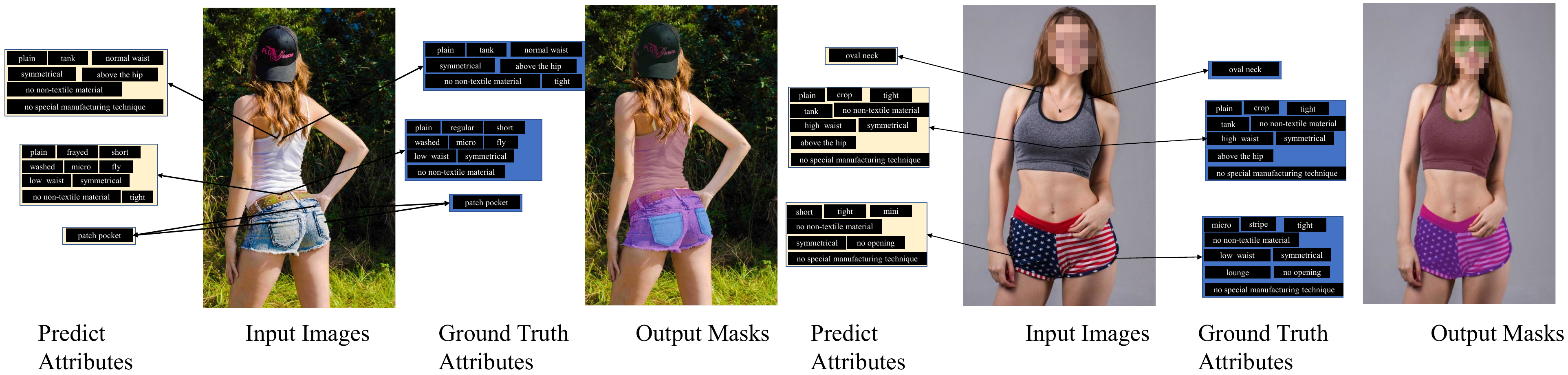}
	\caption{\small Visualization results on Fashionpedia dataset. Our method obtains good segmentation results and almost right attribute predictions. Best view it on screen. }
	\label{fig:vis_fashion_pedia}
\end{figure*}

\begin{figure*}[t!]
	\centering
	\includegraphics[width=0.70\linewidth]{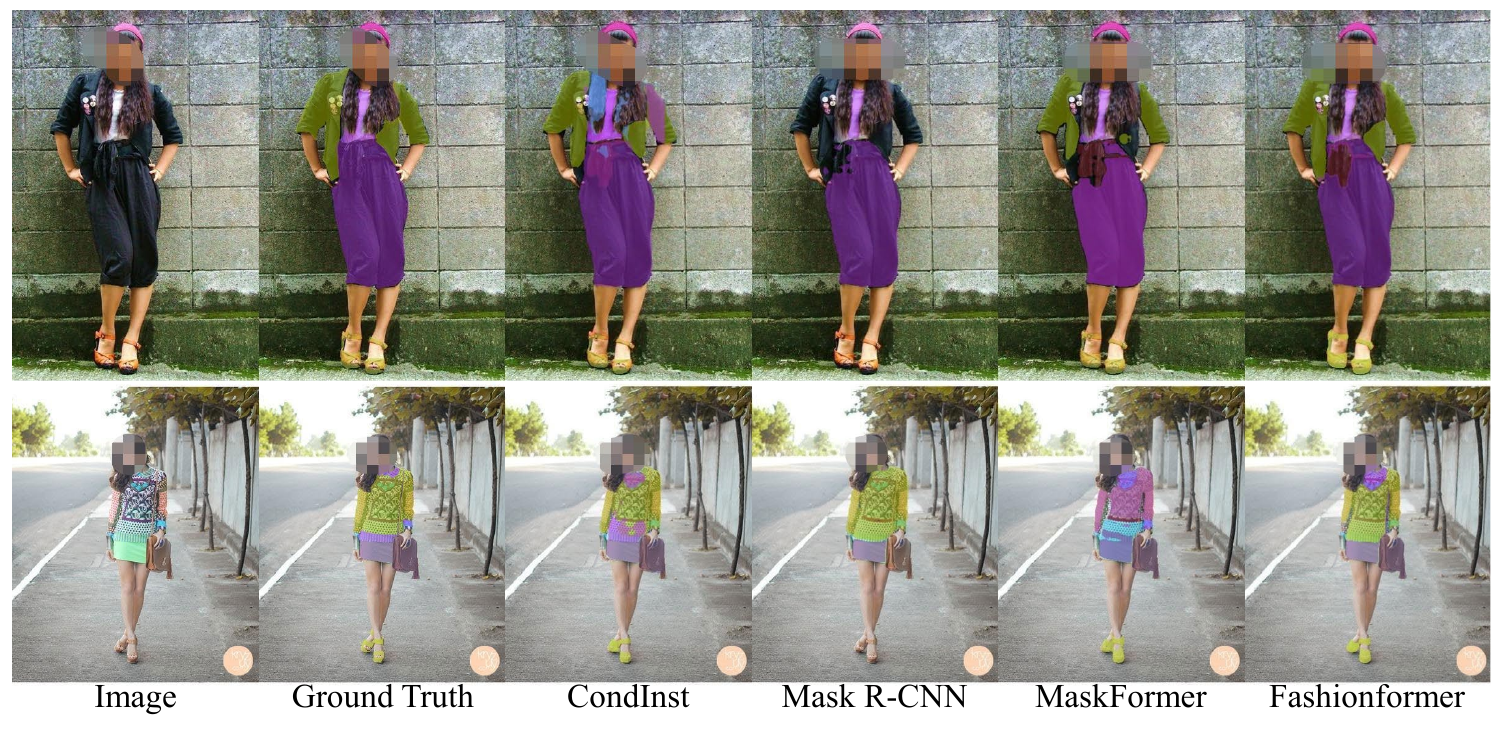}
	\caption{\small Visual comparison results on ModaNet dataset. Our method has better consistent segmentation results. Best view it on screen and zoom in.}
	\label{fig:vis_moda_net}
\end{figure*}

\begin{table}
\begin{center}
\scalebox{0.70}{
\begin{tabular}{ l | c | c | c | c | c | c }
\hline
\textbf{Category}
&\textbf{AP} &\textbf{AP50} &\textbf{AP75} &\textbf{APl} & \textbf{APm} & \textbf{APs}\\
\hline
overall(AM-RCNN)
& 43.1 / 33.3  &60.3 / 42.3 &47.6 / 37.6 &50.0 / 35.4 &40.2 / 27.0 &17.3 / 9.4\\
\hline
outerwear & 64.1 / 40.7 &77.4 / 49.0 &72.9 / 46.2 &67.1 / 43.0 &44.4 / 29.3 &19.0 / 4.4 \\
parts & 19.3 / 13.4 & 35.5 / 20.8 & 18.4 / 14.4 & 28.3 / 14.5 & 23.9 / 16.4 & 12.5 / 9.8 \\
accessory & 56.1 / ~~-~~   & 77.9 / ~~-~~   & 63.9 / ~~-~~   & 57.5 / ~~-~~   & 60.5 / ~~-~~   &~25.0 / ~~-~~   \\
\hline
overall(Fashionformer) & 45.7 / 42.8 & 58.2 / 49.7 & 45.7 / 43.3 & 63.3 / 48.9   & 43.9 / 31.8 & 11.4 /6.5 \\
\hline
outerwear & 72.7 / 52.4 & 78.7 / 56.5 & 73.5 / 53.2 & 76.9 / 55.9 & 48.3 / 31.5 & 8.4 / 2.2 \\ 
parts & 19.2 / 20.3 & 31.8 / 30.1 & 17.2 / 19.7 & 42.6 / 29.0 & 29.8 / 26.2 & 8.2 / 7.1\\ 
accessory & 56.4 / ~~-~~ & 75.0 / ~~-~~ & 58.6 / ~~-~~ & 73.4 / ~~-~~ & 60.9 / ~~-~~ & ~18.1 / ~~-~~ \\
\hline
\end{tabular}
}
\caption{\small Per super-category results. Result format follows [mAP$^{\text{mask}}_{\text{IoU}}$ / mAP$^{\text{mask}}_{\text{IoU+F}_1}$].
AM-RCNN means Attribute-Mask R-CNN with SpineNet-143 backbone. Our results are obtained with ResNet101 backbone. We follow the same COCO sub-metrics for overall and three super-categories for apparel objects. }
\label{tab:per_cls_results}
\end{center}
\end{table}

\noindent
\textbf{Results visualization.} In Fig.~\ref{fig:vis_fashion_pedia}, we present several visualization examples on Fashionpedia using ResNet101 backbone. Our model can predict perfect instance segmentation masks and almost right attributes. Moreover, we compare our model on the ModaNet dataset with other methods. Our model achieves better segmentation results, including more consistent masks and correct labels. More visualization examples can be found in the appendix file.

\noindent
\textbf{Detailed comparison with Attribute-Mask R-CNN.} In Tab.~\ref{tab:per_cls_results}, we show the detailed comparison with  Attribute-Mask R-CNN. Although our backbone is weak (ResNet101 vs. SpineNet-143), our method can achieve better results in most detailed metrics, including AP75, AP50, APl, and APm. However, our method leads to bad results on small objects. The main reason is that our work only considers the single scale feature representation with a simple fusion strategy. In the case of detailed categories, we find that our model achieves better results on ``outerwear'' categories because our method directly outputs masks on a high resolution feature. Moreover, for AP$_{\text{IoU}+\text{F}_1}$, our method achieves better results. For ``part'' categories, our method obtains similar results since most part objects are small. However, our method has \textit{better} attribute recognition results for AP50 and AP75.

\section{Conclusion}
\label{sec:conclusion}
This paper presents a new baseline named Fashionformer for joint fashion segmentation and attribute recognition. Using both object and attribute queries, we present a unified solution in the case of task association for fashion segmentation and attribute recognition. We design a two-stream query update framework in the decoder part with a novel Muti-Layer Rendering module for better attribute recognition. Extensive experiments on the Fashionpedia dataset show the reciprocal benefits on both tasks, where their gap is minimized significantly. We achieve state-of-the-art results on three datasets (Fahsionpedia, ModaNet, and Deepfashion) with fewer GFlops and parameters. Our models also show better results than the recent transformer models in various settings. We hope our method can serve as a new baseline for human fashion understanding.



\bibliographystyle{splncs04}
\bibliography{egbib}

\end{document}